\documentclass{article}

\usepackage{arxiv}

\usepackage{amsmath,amsfonts,bm}

\def\eqref#1{equation~\ref{#1}}
\def\1{\bm{1}}

\DeclareMathAlphabet{\mathsfit}{\encodingdefault}{\sfdefault}{m}{sl}
\SetMathAlphabet{\mathsfit}{bold}{\encodingdefault}{\sfdefault}{bx}{n}

\usepackage[utf8]{inputenc} % allow utf-8 input
\usepackage[T1]{fontenc}    % use 8-bit T1 fonts
\usepackage{hyperref}       % hyperlinks
\usepackage{url}            % simple URL typesetting
\usepackage{booktabs}       % professional-quality tables
\usepackage{amsfonts}       % blackboard math symbols
\usepackage{nicefrac}       % compact symbols for 1/2, etc.
\usepackage{microtype}      % microtypography
\usepackage{lipsum}
\usepackage{graphicx}
\usepackage[table]{xcolor}
\usepackage{multirow}
\usepackage{longtable}
\graphicspath{ {./images/} }
\usepackage{natbib}
\usepackage{subcaption}
\usepackage[ruled,vlined]{algorithm2e}
\usepackage{listings}

\renewcommand{\today}{\vspace{-0.8cm}} % Adjust the -2cm as needed

\title{MARS: Toward More Efficient Multi-Agent Collaboration for LLM Reasoning}

\author{
  Xiao Wang \\
  Indiana University Bloomington \\
  \texttt{xw90@iu.edu}
  \and
  Jia Wang \\
  Indiana University Bloomington \\
  \texttt{jw316@iu.edu}
  \and
  Yijie Wang \\
  Indiana University Bloomington \\
  \texttt{yijwang@iu.edu}
  \and
  Pengtao Dang \\
  Oregon Health \& Science University \\
  \texttt{dangpe@ohsu.edu}
  \and
  Sha Cao \\
  Oregon Health \& Science University \\
  \texttt{caosh@ohsu.edu}
  \and
  Chi Zhang \\
  Oregon Health \& Science University \\
  \texttt{zhangchi@ohsu.edu}
}

\begin{document}
\maketitle
\begin{abstract}
Large language models (LLMs) have achieved impressive results in natural language understanding, yet their reasoning capabilities remain limited when operating as single agents. Multi-Agent Debate (MAD) has been proposed to address this limitation by enabling collaborative reasoning among multiple models in a round-table debate manner. While effective, MAD introduces substantial computational overhead due to the number of agents involved and the frequent communication required.
In this paper, we propose MARS (Multi-Agent Review System), a role-based collaboration framework inspired by the review process. In MARS, an author agent generates an initial solution, reviewer agents provide decisions and comments independently, and a meta-reviewer integrates the feedback to make the final decision and guide further revision. This design enhances reasoning quality while avoiding costly reviewer-to-reviewer interactions, thereby controlling resource consumption.
We compared MARS with both MAD and other state-of-the-art reasoning strategies across multiple benchmarks. Extensive experiments with different LLMs show that MARS matches the accuracy of MAD while reducing token usage by approximately 50\%. Code is available at \url{https://github.com/xwang97/MARS}.
\end{abstract}

\section{Introduction}

Recent advances in large language models (LLMs) have brought revolutionary changes to the field of natural language processing and understanding \citep{minaee2024large, hadi2023large, li2025fundamental, brown2020language}. With their unprecedented performance on language generation tasks such as chat completion, question answering, and code generation, these models are expected to extend their impact to more general and challenging real-world applications, including knowledge-intensive reasoning. However, when it comes to problems involving deep factual knowledge and complex structures, their reasoning abilities still remain limited \citep{mirzadeh2024gsm, jiang2024peek}. For instance, many state-of-the-art models can only achieve about 30\% - 40\% accuracy on the  GPQA benchmark.

To enhance reasoning during inference, several prompting techniques have been proposed, including Chain-of-Thought (CoT) \citep{wei2022chain}, Tree-of-Thought (ToT) \citep{yao2023tree}, and self-reflection \citep{shinn2023reflexion}. While these approaches significantly improve LLM performance, they operate in a single-agent setting, where the model lacks external feedback and often fails to correct its own mistakes. Motivated by this limitation, recent works have explored multi-agent collaboration, in which agents provide feedback to each other to enhance reasoning. Among these, the Multi-Agent Debate (MAD) framework \citep{du2023improving} is the most widely adopted: multiple agents are recruited to tackle the same problem, exchange responses, and refine their answers by considering disagreements, with the final solution derived via majority voting after a round-table discussion.

Despite MAD achieving a significantly improved performance in reasoning, a key drawback is its high resource cost: both token consumption and inference time increase substantially compared to single-agent systems, due to intensive communication among agents. This raises a fundamental question: \textbf{Can we reduce resource consumption while preserving the response quality of multi-agent reasoning?} 

To address this challenge, we draw inspiration from two sources: verifier-based architectures and the hierarchical evaluation structure of peer review. On one hand, recent studies on agent tuning with synthesized trajectories incorporate verifiers or critics as key components \citep{fu2025agentrefine, gao2024multi, shi2025legalreasoner}. These verifiers, either rule-based or LLM-based, check and filter low-quality trajectories or intermediate results, underscoring the importance of systematic review and quality control. On the other hand, peer review practices demonstrate the effectiveness of hierarchical evaluation: authors (or agents) refine their work based on structured feedback, while final decisions are reached through systematic aggregation. Unlike the round-table discussion in MAD, where agents communicate freely, peer review emphasizes independence: reviewers provide evaluations without direct interaction, yet the process reliably converges to fair and consistent outcomes.

Based on these considerations, we propose \textbf{M}ulti-\textbf{A}gent \textbf{R}eview \textbf{S}ystem (\textbf{MARS}) for LLM reasoning, in which agents take on the roles of author, reviewers, and meta-reviewer to mimic a structured review process. Given a user query, the author agent first generates an initial response. Instead of producing their own answers from scratch, reviewer agents evaluate this response, focusing on identifying potential errors. Different to the round-table discussion in MAD, where agents engage in frequent communication, reviewers in MARS operate independently, each providing a decision, review comments, and a confidence score. The meta-reviewer then aggregates these evaluations and issues a final judgment with justification: if the decision is \textit{accept}, the author’s initial response is adopted as the final answer; if the decision is \textit{reject}, the meta-reviewer generates feedback to guide the author’s revision. By structuring collaboration around independent review and feedback, MARS matches the reasoning quality of MAD while consuming substantially fewer tokens and requiring less inference time.

We conducted a series of experiments on three commonly utilized reasoning benchmarks, using both closed-source LLMs (e.g., ChatGPT) and open-source models. With the same number of agents, MARS consistently reduced the resource usage while keeping the accuracy comparable to MAD across all datasets (Table~\ref{tab:pivoted_scores_tokens}). For example, the token consumption of GPT-4o-mini on GPQA is reduced from 17083 to 7903 per query, while achieving a slightly higher accuracy.  These results demonstrated both the effectiveness and efficacy of the MARS framework. In summary, this work has the following major contributions:
\begin{itemize}
    \item We propose \textbf{MARS}, a role-based multi-agent framework by following a hierarchical evaluation process to increase the computational efficiency and preserve the performance of multi-agent reasoning.
    \item Experiments on multiple benchmarks validated that MARS matches the accuracy of the SOTA method MAD while reducing token usage by $\sim$50\% when implementing with different LLMs.
    \item We conduct a comprehensive analysis of multi-agent reasoning dynamics, revealing critical insights into how scaling the number of agents, deploying heterogeneous LLMs across specific roles, and enforcing reviewer personas impact overall system performance.

\end{itemize}

\section{Related Work}
\label{gen_inst}

\textbf{LLM Reasoning}. The reasoning capability of LLMs has become a research focus in recent years, with various methods proposed at both the training and inference stages. Prompting-based approaches, such as CoT\citep{wei2022chain}, ToT\citep{yao2023tree}, and self-reflection\citep{shinn2023reflexion}, introduce structured pipelines to guide the model to decompose the problem and think step-by-step before generating the final answer. These methods are more straightforward and efficient, but the effectiveness is limited by the model's inherent reasoning ability.  In parallel, there are also advanced reasoning models or agents which are finetuned with reasoning trajectories\citep{gao2024multi, tan2025enhancing, huang2022large}. These models come with reasoning and tool usage abilities naturally, but will require more computational resource for model training.

\textbf{Multi-agent Collaboration}. A single agent's capability is often insufficient for problems that require complex reasoning and specialized knowledge. Recent research also revealed that LLMs tend to adhere to their initial answers and struggle to identify and correct their own mistakes without external information. To address this, many studies have started to explore the collaborative behaviors of multiple agents. MAD is the most prominent framework in this line\citep{du2023improving, smit2023should, liang2023encouraging}, where multiple agents work on the same problem through round-table discussion, iteratively exchanging responses and refining their answers by leveraging the feedback from others. While effective in improving reasoning, MAD introduces substantial computational overhead: both token consumption and inference time grow rapidly with the number of agents and communication rounds, making it costly for practical deployment.

\textbf{Answer Verification}. Many finetuning methods for LLM agents rely on synthetic reasoning and tool usage trajectories\citep{sun2025mm, ficek2025scoring, lee2025revise, fu2025agentrefine}. To ensure the quality of the structure, intermediate steps, and final results of the trajectories, recent works have designed various verifiers for quality control. These verifiers are either rule-based modules with human-defined principles, or LLM-based utilizing the inherent error-detection and evaluation abilities of large models. Unlike primary agents, verifiers are not required to generate full answers to user queries; instead, they specialize in detecting mistakes in reasoning steps and final results. The effectiveness of this design in improving reliability has been demonstrated in various agent systems. In our framework, reviewer agents serve a similar role: they provide independent decisions, comments, and suggestions, offering the author agent clearer and more explicit feedback to guide revisions of its initial response.

\section{Methodology}
\label{method}

\subsection{Overview}
In this section, we introduce the MARS framework, which is designed for reducing resource demands in multi-agent systems. We begin with the problem formulation, then describe the roles assigned to different agents, and finally outline the architecture and mechanisms of how the MARS framework works for complex reasoning tasks.

\textbf{Problem Formulation.} For a given reasoning task, let $x_i$ denote the input prompt (original problem description wrapped by our prompting templates), and let $y_i$ denote the model's response containing the final answer. Different from MAD where multiple agents are equivalent to each other, MARS adopts a role-based design. LLMs are employed to play three distinct roles, namely \textbf{author}, who generates initial responses, \textbf{reviewer}, who generates evaluation on the author's output, and \textbf{meta-reviewer}, who integrates reviewers' comments and provides feedback for the author. We use $\mathcal{A}$ to denote the author agent, $\mathcal{R}_j, j=1,2,...,m$ to denote the reviewer agents, and $\mathcal{M}$ to denote the meta-reviewer agent, respectively.

\subsection{MARS Framework}

\begin{figure*}
    \centering
    \includegraphics[width=1\linewidth]{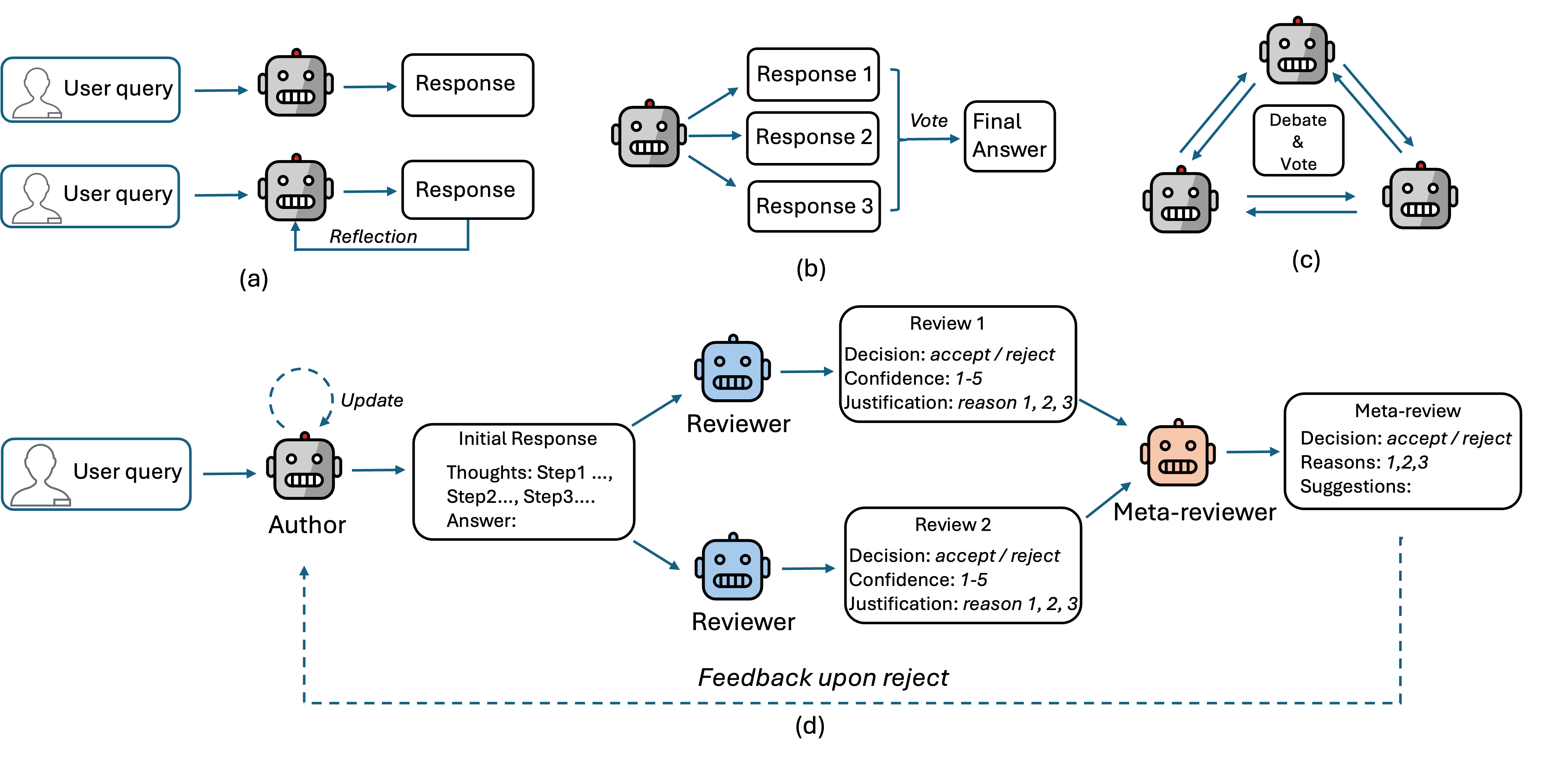}
    \caption{Overview of the architecture of MARS and baselines. (a) Basic single model inference (top) and self-reflection (bottom). (b) Self-consistency. (c) Multi-Agent-Debate (MAD). (d) Multi-Agent Review System (MARS). In MARS, the author agent receives a user query and generates an initial response. Each reviewer agent evaluates the response and provides a decision, confidence level, and justification (e.g., reasons for the decision, identified author mistakes). The meta-reviewer integrates review comments and makes the final decision, with suggestions for answer revision. Finally, the author agent incorporates the feedback and updates its response, leading to enhanced reasoning.}
    \label{fig:framework}
\end{figure*}

The MARS framework implements a structured, role-based collaboration among agents to improve both reasoning accuracy and efficiency (Figure~\ref{fig:framework}). Given a user query, the author agent generates an initial response. This output is then independently evaluated by multiple reviewer agents, each producing a decision, explanatory comments, and a confidence score without interacting with each other. A meta-reviewer agent aggregates these reviews to issue a judgment and, when necessary, provides targeted feedback to guide the author’s revision. This hierarchical evaluation and feedback loop enables efficient parallel review while enhancing robustness through systematic error detection and correction. The details of each component and step are provided below (a summary of the procedures is available in Algorithm~\ref{alg:mars}).

\textbf{Author's initial response.} For an input prompt $x_i$, the \textbf{author agent} generates an initial response containing both the reasoning trajectory and the final answer. To make the subsequent review process verifiable, we adopt the Chain-of-Thought (CoT) paradigm, requiring the author to output explicit intermediate reasoning steps alongside the final answer (see Appendix~\ref{author_prompt} for the prompt template). Mathematically, this step can be presented as a function:
\begin{equation}
\label{eq_author}
    t_i, y_i = \mathcal{A}(x_i)
\end{equation}
where $t_i$ denotes the sequence of reasoning steps and $y_i$ denotes the final answer.

\textbf{Reviewer's evaluation.} After the author produces an initial response, the review stage begins with $m$ \textbf{reviewer agents}, denoted $\mathcal{R}_1, \mathcal{R}_2, \ldots, \mathcal{R}_m$. Each reviewer receives the author’s output as input and generates a review comprising: (i) a binary decision (accept or reject), (ii) a confidence score, and (iii) textual justification. If the decision is \textit{reject}, the justification must explicitly identify errors in either the intermediate reasoning steps or the final prediction (see Appendix~\ref{reviewer_prompt} for the prompt template). The review process of the $j$-th reviewer for the input $x_i$ can be mathematically presented as a function:
\begin{equation}
\label{eq_review}
    r_{ij} = \mathcal{R}_j(x_i, t_i, y_i), \quad j=1,2,...,m
\end{equation}

\textbf{Meta-review and feedback.} Unlike the normal peer review process, we do not incorporate direct feedback from the reviewers to the author after the review stage. Instead, a \textbf{meta-reviewer agent} consolidates reviewer feedback, makes decisions, and communicates with the author. This design is driven by two factors: (1) \textbf{Efficiency} — a direct discussion between reviewers and the author would substantially increase token usage and inference time, reducing reasoning efficiency; (2) \textbf{Role simplification} — Exposing the author to multiple, potentially redundant or conflicting comments would impose extra decision-making, such as filtering overlapping feedback or resolving disagreements, which could distract it from revising the response. We consider that by centralizing feedback integration in the meta-reviewer, the author remains focused on updating their answer, thereby maintaining reasoning quality.

The meta-reviewer processes all reviewer outputs, summarizes comments, resolves conflicts or redundancies, and issues a final decision. If the decision is \textit{accepted}, the author’s initial response is adopted as the final response. If the decision is \textit{rejected}, the meta-reviewer provides explicit feedback to the author, including (i) the reasons for rejection (e.g., errors in the reasoning path) and (ii) concrete suggestions for improvement (see Appendix~\ref{meta_prompt} for the prompt template). This meta-reviewer can be mathematically presented as a function:
\begin{equation}
\label{eq_meta}
    m_i = \mathcal{M}(x_i, t_i, y_i, r_i)
\end{equation}
, where $m_i$ denotes the meta-reviewer's output, and $r_i$ is a concatenation of all the reviewer outputs, $r_i = r_{i1} \oplus r_{i2} \oplus ... \oplus r_{im}$

\textbf{Rebuttal stage.} After the meta-review stage, a final decision is issued regarding the author’s initial response. If the decision is \textit{rejected}, the author agent is required to revise the solution by leveraging both its reasoning history and the meta-reviewer’s feedback. Concretely, the author agent will review its step-by-step thoughts $t_i$ and incorporate the meta-reviewer’s suggestions to identify and correct errors (see Appendix~\ref{feedback_prompt} for the prompt template). Once all steps have been re-examined, the author agent produces an updated answer $y_i^*$. This process can be formulated as:
\begin{equation}
\label{eq_rebuttal}
    y_i^* = \mathcal{A}(t_i, y_i, m_i)
\end{equation}
Upon generating $y_i^*$, the updated response is fed back into the review stage. This iterative cycle of review and revision continues until the meta-reviewer issues an \textit{accepted} decision or a pre-defined maximum number of iterations $K$ is reached (see Algorithm~\ref{alg:mars}).

\subsection{Calibration of Reviewer Confidence}

Reviewers may occasionally make reasoning errors, leading to divergent decisions and uncertainties at the review stage. To assist the meta-reviewer in accurately weighing these evaluations, it is essential to quantify reviewer confidence. As detailed previously, our framework prompts each reviewer to output a verbalized confidence score alongside their decision and justification. However, recent studies demonstrate that LLMs frequently exhibit significant overconfidence when relying on vanilla verbalized confidence elicited via prompting \citep{xiong2023can}. 

Conversely, research indicates that the probability distribution of output tokens intrinsically reflects a model's underlying uncertainty. Building on this insight, we propose utilizing the token-level probability distribution of the reviewers' responses for confidence calibration. Specifically, for a response $r_{ij}$ generated by the $j$-th reviewer for the $i$-th query, consisting of $N$ tokens $\{t_1, t_2, \dots, t_N\}$, we first compute the average log probability of the generated sequence:
\begin{equation}
    \text{AvgLogProb}_{ij} = \frac{1}{N} \sum_{k=1}^{N} \log P(t_k)
\end{equation}

A higher average log probability (closer to 0) indicates greater model certainty. To convert this into a normalized confidence score bounded between 0 and 1, we exponentiate the average log probability:
\begin{equation}
    Conf_{ij} = \exp(\text{AvgLogProb}_{ij})
\end{equation}

In practice, this sequence-level probability calibrates the initial confidence level of each reviewer. This refined metric is then propagated to the meta-reviewer via the prompt template (\ref{confidence}), providing a statistically grounded indicator of whether a given reviewer's feedback is reliable.

\section{Experiments}
\subsection{Setup}
\label{setup}
\textbf{Benchmarks.} We evaluate MARS on three commonly utilized reasoning benchmarking sets: (1) MMLU \citep{hendrycks2020measuring} consists of multiple-choice questions across diverse academic subjects such as elementary mathematics, history, computer science, and law; (2) GPQA \citep{rein2024gpqa} is a graduate-level QA dataset spanning biology, physics, and chemistry, which is of high-quality and even challenging for domain experts, making it a strong test of reasoning ability; (3) GSM8K \citep{cobbe2021training} is a math word problem dataset that evaluates grade school level mathematical reasoning.

\textbf{Models and baselines.} We use both closed-source (GPT-3.5-turbo, GPT-4o-mini) and open-source (Mixtral-8$\times$7b, Mixtral-8$\times$22b, Llama3.3-70b) models as agent backbones. The open-source model is accessed through NVIDIA NIM platform via API calls. We compare MARS against several single-agent and multi-agent baselines, including Chain-of-Thought (CoT) \citep{wei2022chain}, Self-reflection \citep{shinn2023reflexion}, Self-consistency \citep{wang2022self}, Multi-agent Debate (MAD) \citep{du2023improving}, and Diverse Multi-agent Debate (DMAD) \citep{liu2025breaking}. Detailed introductions and experimental settings of the baselines, as well as their corresponding prompts are available in Appendix~\ref{ex_details} and Appendix~\ref{all_prompts}.

\textbf{Metrics.} Prior work primarily assessed reasoning methods by the accuracy of the final answer, where MAD has demonstrated strong performance. In contrast, MARS is designed to improve computational efficiency in LLM-based reasoning, with success defined by achieving accuracy comparable to MAD while substantially reducing resource consumption. Accordingly, we evaluate methods based on the integration of three metrics: (i) final-answer accuracy, (ii) average number of tokens generated per query, and (iii) average inference time (seconds) per query.

\subsection{Main Results}
\label{mainresults}

\begin{table*}[t]
% \caption{Performance comparison across datasets including Accuracy (Acc. \%) and average number of tokens (Tok.). Best accuracy scores per dataset are \textbf{bolded}, and second-best are \underline{underlined}.}
\centering
\setlength{\tabcolsep}{6pt} % Reduced to fit all 11 columns
\renewcommand{\arraystretch}{1.1}
\small % Adjusting font size to fit the page width
\begin{tabular}{l cc cc cc cc cc}
\toprule
\multirow{2}{*}{Method} & \multicolumn{2}{c}{GPT-3.5} & \multicolumn{2}{c}{GPT-4o-mini} & \multicolumn{2}{c}{Mixtral 8x7b} & \multicolumn{2}{c}{Mixtral 8x22b} & \multicolumn{2}{c}{Llama3-70b} \\
\cmidrule(lr){2-3} \cmidrule(lr){4-5} \cmidrule(lr){6-7} \cmidrule(lr){8-9} \cmidrule(lr){10-11}
& Acc. & Tok. & Acc. & Tok. & Acc. & Tok. & Acc. & Tok. & Acc. & Tok. \\
\midrule
\multicolumn{11}{c}{\textbf{GPQA}} \\
CoT              & 31.00 & 414 & 43.33 & 677 & 33.67 & 759 & 41.33 & 721 & 47.67 & 963 \\
Self-reflection  & 29.67 & 890 & 37.67 & 1597 & 35.00 & 1625 & 38.00 & 1579 & 56.33 & 2079 \\
Self-consistency & 33.67 & 1686 & 40.67 & 2910 & 35.67 & 3033 & 41.67 & 2840 & 56.67 & 3470 \\
MAD              & 31.00 & 7567 & \underline{47.50} & 17083 & \underline{35.67} & 12835 & \textbf{47.00} & 13308 & \underline{59.67} & 19255 \\
DMAD              & \underline{35.67} & 5609 & 47.33 & 17229 & 34.00 & 14139 & 43.33 & 13509 & 56.33 & 18082 \\
MARS             & \textbf{36.33} & 3741 & \textbf{48.33} & 7903 & \textbf{37.33} & 6441 & \underline{44.00} & 5418 & \textbf{60.00} & 6633 \\
\midrule
\multicolumn{11}{c}{\textbf{MMLU}} \\
CoT              & 68.00 & 283 & 81.67 & 357 & \underline{62.00} & 438 & 73.33 & 434 & 84.00 & 446 \\
Self-reflection  & 70.33 & 609 & 80.67 & 847 & 60.00 & 1042 & 77.33 & 988 & 83.67 & 1087 \\
Self-consistency & 69.33 & 1141 & 85.00 & 1515 & 61.67 & 1742 & 76.33 & 1757 & 82.33 & 1919 \\
MAD              & \textbf{71.33} & 4647 & \underline{85.33} & 8338 & 61.00 & 7118 & 76.67 & 7200 & 84.00 & 9370 \\
DMAD              & 71.00 & 4023 & 85.33 & 8706 & 54.33 & 8658 & \textbf{83.00} & 7570 & \underline{84.67} & 9641 \\
MARS             & \underline{71.00} & 2692 & \textbf{85.67} & 4210 & \textbf{63.00} & 3887 & \underline{77.67} & 3829 & \textbf{84.67} & 4210 \\
\midrule
\multicolumn{11}{c}{\textbf{GSM8K}} \\
CoT              & 73.33 & 262 & 96.00 & 384 & 60.00 & 381 & 86.33 & 373 & 97.00 & 335 \\
Self-reflection  & 67.00 & 577 & 92.33 & 894 & 53.67 & 826 & 86.00 & 817 & 97.33 & 777 \\
Self-consistency & \underline{77.00} & 1045 & 97.00 & 1548 & 67.00 & 1536 & 87.00 & 1517 & 97.33 & 1410 \\
MAD              & \textbf{79.00} & 4355 & \underline{97.67} & 7440 & \underline{68.67} & 6809 & \underline{87.00} & 6805 & \underline{97.33} & 6307 \\
DMAD              & 73.33 & 5136 & 96.33 & 8831 & 66.33 & 7724 & 83.33 & 7283 & 97.33 & 7659 \\
MARS             & 75.67 & 2508 & \textbf{98.00} & 3389 & \textbf{69.00} & 4994 & \textbf{90.33} & 3515 & \textbf{97.67} & 3313 \\
\bottomrule
\end{tabular}
\caption{Performance comparison across datasets including Accuracy (Acc. \%) and average number of tokens (Tok.). Best accuracy scores per dataset are \textbf{bolded}, and second-best are \underline{underlined}.}
\label{tab:pivoted_scores_tokens}
\end{table*}

The primary experimental results across all models and datasets are summarized in Table~\ref{tab:pivoted_scores_tokens}. We highlight the following key observations:

(1) \textbf{Accuracy.} MAD and MARS consistently rank among the top two methods in final-answer accuracy, outperforming all of the other baselines under most settings. With the expanded evaluation across five diverse LLMs, MARS demonstrates superior or highly competitive performance. Their relative advantage depends on the backbone model and task: for example, on GSM8K with GPT-3.5, MAD reaches 79.00\% accuracy versus 75.67\% for MARS. However, MARS outperforms MAD on the remaining four models for the same task, including achieving 90.33\% with Mixtral 8x22b (compared to MAD's 87.00\%). Furthermore, MARS secures the highest accuracy on the challenging GPQA dataset with advanced models like GPT-4o-mini (48.33\%) and Llama3-70b (60.00\%). Overall, MARS achieves comparable or superior accuracy to MAD across an extensive array of state-of-the-art models and complex tasks.

(2) \textbf{Token consumption.} Although both MAD and MARS achieve high accuracy across reasoning tasks, their multi-agent design naturally requires higher token usage compared to single-agent baselines. Notably, MARS consistently reduces the average number of tokens per query by roughly 50\% relative to MAD, a scaling advantage that holds true across all tested models. For instance, on the GPQA benchmark with GPT-4o-mini, MAD consumes an average of 17,083 tokens per query, whereas MARS requires only 7,903 tokens. Given that MARS achieves comparable or better accuracy, this demonstrates that the framework makes the multi-agent collaboration paradigm substantially more efficient and practical for large-scale deployment while preserving response quality.

(3) \textbf{Inference time.} Another critical aspect of a reasoning strategy is inference time, particularly for interactive applications such as ChatGPT. As shown in Table~\ref{tab:inference_time}, MARS reduces inference time by more than 30\% compared to MAD, enabling more responsive user interactions and more scalable large-scale experimentation.

\begin{figure}[h]
    \centering
    \includegraphics[width=\linewidth]{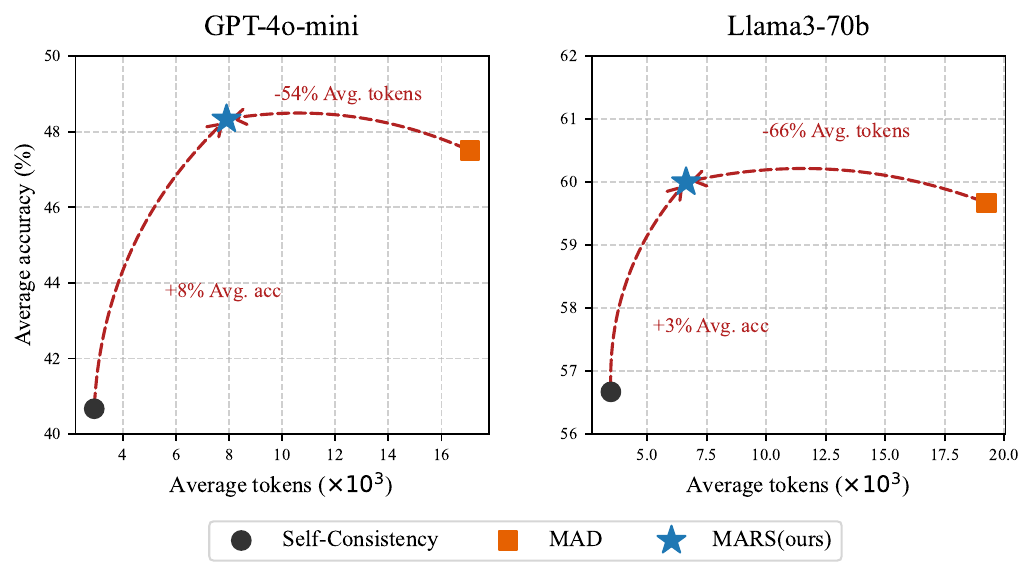}
    \caption{Accuracy-Resource trade-off across different models on GPQA. MARS demonstrates a significant reduction in token averaged tokens compared to MAD, while achieving higher accuracy than self-consistency.}
    \label{fig:tradeoff}
\end{figure}

Our results reveal a key limitation of prior MAD-inspired strategies: their gains in reasoning quality come with prohibitive computational overhead. MARS fills this efficiency gap to a certain level, reducing token usage and inference time by about 50\% while preserving accuracy. This advance makes multi-agent reasoning not only more effective but also practical for large-scale and real-time applications. Figure~\ref{fig:tradeoff} visualizes this efficiency gain, highlighting the favorable trade-off MARS achieves between accuracy and token consumption relative to self-consistency and MAD.
% \vspace{-1cm}

\begin{figure}[h]
    \centering
    % First row: GPT
    \begin{subfigure}{0.49\linewidth}
        \centering
        \includegraphics[width=\linewidth]{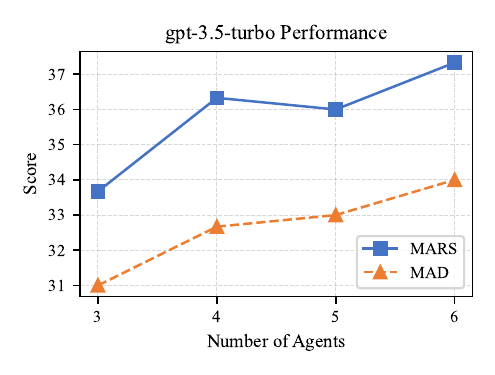}
        % \caption{Performance trend}
        \label{fig:score}
    \end{subfigure}
    \hfill
    \begin{subfigure}{0.49\linewidth}
        \centering
        \includegraphics[width=\linewidth]{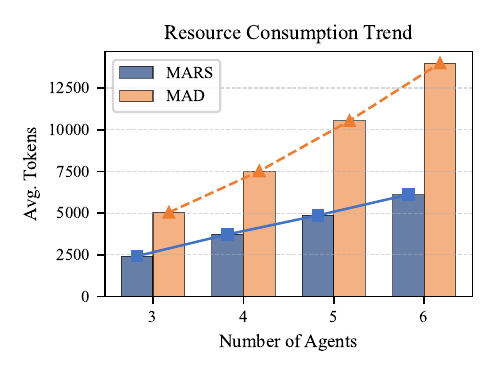}
        % \caption{Token consumption}
        \label{fig:tokens}
    \end{subfigure}
    
    \vspace{-0.3cm} % Optional: adds a bit of vertical space between the rows
    
    % Second row: Mixtral
    \begin{subfigure}{0.49\linewidth}
        \centering
        \includegraphics[width=\linewidth]{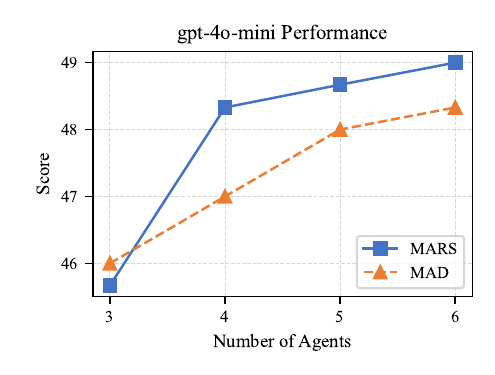}
        % \caption{Performance trend}
        \label{fig:score2}
    \end{subfigure}
    \hfill
    \begin{subfigure}{0.49\linewidth}
        \centering
        \includegraphics[width=\linewidth]{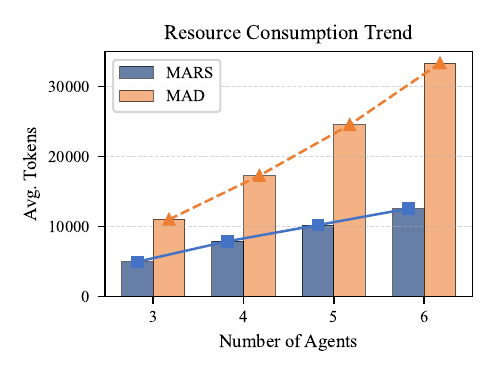}
        % \caption{Token consumption}
        \label{fig:tokens2}
    \end{subfigure}
    \vspace{-1cm}
    \caption{Comparison of MARS and MAD on GPQA with varying number of agents. Row 1: using GPT-3.5-turbo as the backbone; Row 2: using GPT-4o-mini as the backbone. Column 1: accuracy scores; Column 2: averaged number of tokens.}
    \label{fig:scale}
\end{figure}

% \vspace{-0.1cm}
\subsection{Additional Analysis}
\subsubsection{Scalability}
In this section, we evaluate the scalability of MARS versus MAD to further demonstrate the efficiency advantage of MARS. Figure~\ref{fig:scale} presents their performance across different backbone models and varying numbers of agents. From the left column of Figure~\ref{fig:scale}, we observe that the accuracy of both methods generally increases as the number of agents grows, consistent with prior findings that collaborative reasoning benefits from larger agent pools. This trend suggests that scaling the number of agents can enhance reasoning performance in multi-agent frameworks.
Noted, MAD employs a fully connected communication structure that causes computational cost per query to grow rapidly as the number of agents increases. As shown in the right column of Figure~\ref{fig:scale}, both token usage and inference time grow sharply with additional debaters, severely limiting scalability. In contrast, MARS avoids this bottleneck by designing the parallel reviewer to generate independent evaluations and a meta-reviewer agent to aggregate reviews. This structure enables MARS to improve reasoning efficiency while maintaining performance, with token consumption and inference time growing linearly with the number of reviewers.

\subsubsection{Evaluating Different Mixtures of Models}

In this section, we evaluate how utilizing heterogeneous backbone models impacts reasoning outcomes. While our primary experiments deployed the same LLM across all agent roles, recent studies suggest that mixing different models can introduce complementary analytical perspectives. To explore this, we tested various combinations of GPT-3.5-turbo and Mixtral 8x22b within the MARS framework on the GPQA benchmark. As shown in Table~\ref{tab:mixture}, three key observations emerge: (1) \textbf{Stronger auxiliary agents elevate performance.} For a GPT author, integrating the more capable Mixtral model into the reviewer or meta-reviewer roles consistently improves the final score, scaling from 35.33 up to 39.00. (2) \textbf{The author's inherent capacity bounds the system.} Despite the assistance of strong Mixtral reviewers and meta-reviewers, a GPT author (maximum 39.00) cannot match the baseline performance of a Mixtral author (minimum 41.67). The author agent ultimately dictates the overall performance ceiling. (3) \textbf{Model diversity yields unexpected synergies.} The highest overall accuracy (46.40) is achieved when Mixtral serves as the author, but GPT acts as both the reviewers and the meta-reviewer. This configuration even surpasses all-Mixtral settings, indicating that injecting diverse reasoning styles during the review phase is a non-trivial factor that warrants further investigation.
\begin{table}[h]
% \caption{Comparison of different author–reviewer–meta-reviewer configurations on GPQA.}
\centering
\setlength{\tabcolsep}{15pt}
\renewcommand{\arraystretch}{1.0}
\begin{tabular}{l l l c}
\toprule
Author & Reviewers & Meta & Acc.  \\
\midrule
gpt & gpt+mixtral      & gpt      & 35.33 \\
gpt & mixtral+mixtral  & gpt      & 36.00 \\
gpt & gpt+mixtral      & mixtral  & 38.00 \\
gpt & mixtral+mixtral  & mixtral  & 39.00 \\
\midrule
mixtral & gpt+mixtral  & mixtral  & 43.50 \\
mixtral & gpt+gpt      & mixtral  & 41.67 \\
mixtral & gpt+mixtral  & gpt      & 42.23 \\
mixtral & gpt+gpt      & gpt      & 46.40 \\
\bottomrule
\end{tabular}
\caption{Comparison of different author–reviewer–meta-reviewer configurations on GPQA.}
\label{tab:mixture}
\end{table}

\subsubsection{Effect of reviewer personas}
Recent work suggests that assigning distinct personas to agents can improve performance by promoting diversity. Motivated by this, we explored whether diversifying reviewers through predefined personas could enhance feedback quality and overall reasoning performance. In this variant, namely MARS-P, reviewers were instructed to adopt specific roles (e.g., conservative vs. aggressive; see Appendix~\ref{persona_prompt} for prompt details) when evaluating the author’s response.

Surprisingly, as shown in Table~\ref{tab:persona_ablation}, introducing personas did not yield improvements across tasks (except a marginal improvement on MMLU). Analysis of the review content suggests two main factors: (i) the inherent randomness of LLMs already provides a degree of diversity in reviewer feedback, and persona prompts did not add significant variance; and (ii) when the author’s response is correct, aggressive reviewers often generate unnecessarily critical or noisy feedback, which can confuse the meta-reviewer and cause over-corrections. These results indicate that relying on the model’s natural variability might be more effective than enforcing artificial personas.

\begin{table}[h]
% \caption{Ablation study on persona assignment across datasets. Scores are reported as (w/o Persona / w/ Persona).}
\centering
\small % Adjust font size slightly if needed to guarantee a perfect fit
\setlength{\tabcolsep}{10pt} % Tighten column spacing for single-column fit
\renewcommand{\arraystretch}{1.2}
\begin{tabular}{l ccc}
\toprule
Model & GPQA & MMLU & GSM8K \\
\midrule
GPT-3.5        & 36.33 / 34.00 & 71.00 / 72.00 & 75.67 / 73.33 \\
GPT-4o-mini    & 48.33 / 47.00 & 85.67 / 85.67 & 98.00 / 97.00 \\
Mixtral 8x7b   & 37.33 / 35.67 & 63.00 / 66.67 & 69.00 / 66.00 \\
Mixtral 8x22b  & 44.00 / 40.00 & 77.67 / 80.00 & 90.33 / 88.67 \\
Llama3.3-70b   & 60.00 / 56.67 & 84.67 / 82.67 & 97.67 / 97.00 \\
\bottomrule
\end{tabular}
\caption{Ablation study on persona assignment across datasets. Scores are reported as (w/o Persona / w/ Persona).}
\label{tab:persona_ablation}
\end{table}

\subsection{Case Study}
To illustrate how MARS operates, we present a case study from the GSM dataset that demonstrates the framework’s step-by-step reasoning process (Figure~\ref{fig:maincase}; additional examples appear in Appendix~\ref{append_case}). The problem describes four dogs, where the first weighs 10 pounds, and each subsequent dog weighs a constant multiple of the previous one. The task is to determine the weight of the fourth dog.

The author agent began by denoting the first dog’s weight as $A$ and correctly deriving that the fourth weight should be $22A$. However, it mistakenly recomputed $A$ as $5/11$, which contradicts the given information and led to an incorrect final answer. During the review phase, Reviewer 1 successfully identified this error and explicitly flagged the inconsistency; Reviewer 2 also rejected the author’s answer but failed to provide a clear justification. At the meta-review stage, the meta-reviewer integrated these signals, recognized the erroneous recomputation of $A$, and instructed the author to recalculate using the correct initial weight. Incorporating this feedback, the author substituted $A = 10$ pounds and derived the correct result of $22A = 220$ pounds.

We highlight three insights from this case: (1) \textit{Error propagation in long reasoning chains.} For tasks requiring multiple reasoning steps (seven in this example), even a minor mistake can cascade into an incorrect final answer, underscoring the importance of intermediate verifications for the reasoning process. (2) \textit{Reviewer contributions.} Reviewers in MARS can identify errors in the reasoning chain and provide detailed feedback to guide answer revision. While not all reviewers generate high-quality comments, this variability suggests potential benefits from involving multiple reviewers. (3) \textit{Effectiveness of MARS}. By incorporating external feedback rather than relying solely on self-correction, MARS overcomes limitations of single-agent reasoning and improves final accuracy without incurring excessive computational cost.

\begin{figure}[h]
    \centering
    \includegraphics[width=1\linewidth]{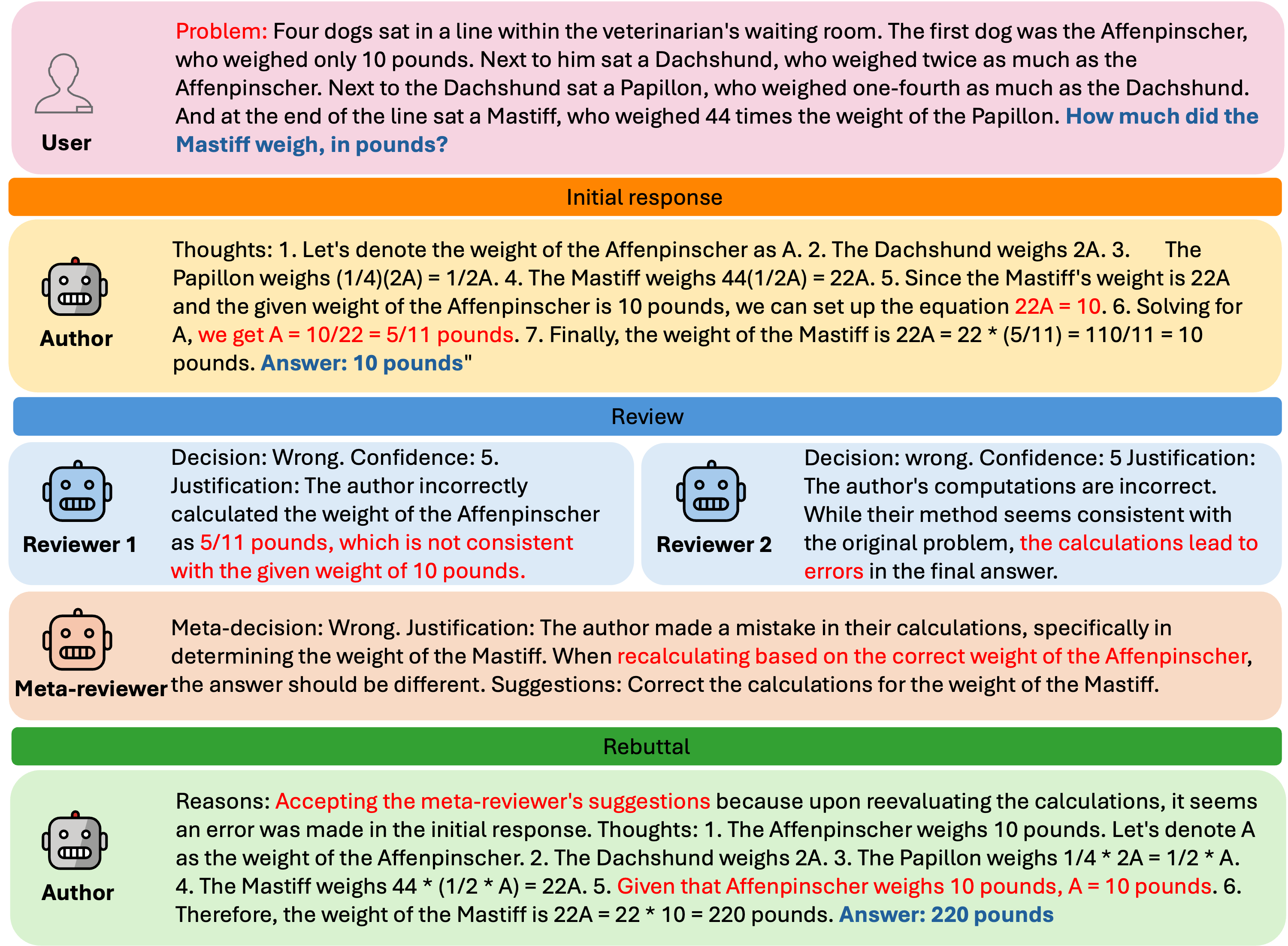}
    \caption{Case study of MARS on a GSM example. Upon receiving the user query, the author agent first generates an initial response but incorrectly recomputed a given variable, leading to an incorrect final answer. The reviewers identified the mistake and generated feedback, which guided the author agent to revise the solution, resulting in the correct answer.}
    \label{fig:maincase}
\end{figure}

\vspace{-0.5cm}
\section{Conclusion}
In this work, we introduced MARS (Multi-Agent Review System), a framework that enables efficient collaboration among LLMs through a propose–review–feedback–update architecture. MARS addresses the limitations of self-reflection in single-agent settings, improves reasoning quality, and avoids the heavy computational overhead of debate-based approaches. Extensive experiments across diverse benchmarks and models demonstrate that MARS consistently outperforms strong baselines and achieves the same level of accuracy with Multi-Agent Debate, while substantially reducing resource consumption. These results suggest that direct communication among all agents is unnecessary for effective collaboration, and that review-style architectures offer a promising new paradigm for efficient multi-agent reasoning. Limitations and future work are discussed in Appendix~\ref{limitations}.


% \bibliography{iclr2025_conference}
% \bibliographystyle{iclr2025_conference}

\clearpage
\bibliographystyle{plain}
\bibliography{references}

@article{minaee2024large,
  title={Large language models: A survey},
  author={Minaee, Shervin and Mikolov, Tomas and Nikzad, Narjes and Chenaghlu, Meysam and Socher, Richard and Amatriain, Xavier and Gao, Jianfeng},
  journal={arXiv preprint arXiv:2402.06196},
  year={2024}
}

@article{hadi2023large,
  title={Large language models: a comprehensive survey of its applications, challenges, limitations, and future prospects},
  author={Hadi, Muhammad Usman and Qureshi, Rizwan and Shah, Abbas and Irfan, Muhammad and Zafar, Anas and Shaikh, Muhammad Bilal and Akhtar, Naveed and Wu, Jia and Mirjalili, Seyedali and others},
  journal={Authorea preprints},
  volume={1},
  number={3},
  pages={1--26},
  year={2023},
  publisher={Authorea}
}

@article{li2025fundamental,
  title={Fundamental capabilities and applications of large language models: A survey},
  author={Li, Jiawei and Gao, Yang and Yang, Yizhe and Bai, Yu and Zhou, Xiaofeng and Li, Yinghao and Sun, Huashan and Liu, Yuhang and Si, Xingpeng and Ye, Yuhao and others},
  journal={ACM Computing Surveys},
  year={2025},
  publisher={ACM New York, NY}
}

@article{brown2020language,
  title={Language models are few-shot learners},
  author={Brown, Tom and Mann, Benjamin and Ryder, Nick and Subbiah, Melanie and Kaplan, Jared D and Dhariwal, Prafulla and Neelakantan, Arvind and Shyam, Pranav and Sastry, Girish and Askell, Amanda and others},
  journal={Advances in neural information processing systems},
  volume={33},
  pages={1877--1901},
  year={2020}
}

@article{mirzadeh2024gsm,
  title={Gsm-symbolic: Understanding the limitations of mathematical reasoning in large language models},
  author={Mirzadeh, Iman and Alizadeh, Keivan and Shahrokhi, Hooman and Tuzel, Oncel and Bengio, Samy and Farajtabar, Mehrdad},
  journal={arXiv preprint arXiv:2410.05229},
  year={2024}
}

@article{jiang2024peek,
  title={A peek into token bias: Large language models are not yet genuine reasoners},
  author={Jiang, Bowen and Xie, Yangxinyu and Hao, Zhuoqun and Wang, Xiaomeng and Mallick, Tanwi and Su, Weijie J and Taylor, Camillo J and Roth, Dan},
  journal={arXiv preprint arXiv:2406.11050},
  year={2024}
}

@article{wei2022chain,
  title={Chain-of-thought prompting elicits reasoning in large language models},
  author={Wei, Jason and Wang, Xuezhi and Schuurmans, Dale and Bosma, Maarten and Xia, Fei and Chi, Ed and Le, Quoc V and Zhou, Denny and others},
  journal={Advances in neural information processing systems},
  volume={35},
  pages={24824--24837},
  year={2022}
}

@article{yao2023tree,
  title={Tree of thoughts: Deliberate problem solving with large language models},
  author={Yao, Shunyu and Yu, Dian and Zhao, Jeffrey and Shafran, Izhak and Griffiths, Tom and Cao, Yuan and Narasimhan, Karthik},
  journal={Advances in neural information processing systems},
  volume={36},
  pages={11809--11822},
  year={2023}
}

@article{shinn2023reflexion,
  title={Reflexion: Language agents with verbal reinforcement learning},
  author={Shinn, Noah and Cassano, Federico and Gopinath, Ashwin and Narasimhan, Karthik and Yao, Shunyu},
  journal={Advances in Neural Information Processing Systems},
  volume={36},
  pages={8634--8652},
  year={2023}
}

@inproceedings{du2023improving,
  title={Improving factuality and reasoning in language models through multiagent debate},
  author={Du, Yilun and Li, Shuang and Torralba, Antonio and Tenenbaum, Joshua B and Mordatch, Igor},
  booktitle={Forty-first International Conference on Machine Learning},
  year={2023}
}

@article{fu2025agentrefine,
  title={Agentrefine: Enhancing agent generalization through refinement tuning},
  author={Fu, Dayuan and He, Keqing and Wang, Yejie and Hong, Wentao and Gongque, Zhuoma and Zeng, Weihao and Wang, Wei and Wang, Jingang and Cai, Xunliang and Xu, Weiran},
  journal={arXiv preprint arXiv:2501.01702},
  year={2025}
}

@article{gao2024multi,
  title={Multi-modal agent tuning: Building a vlm-driven agent for efficient tool usage},
  author={Gao, Zhi and Zhang, Bofei and Li, Pengxiang and Ma, Xiaojian and Yuan, Tao and Fan, Yue and Wu, Yuwei and Jia, Yunde and Zhu, Song-Chun and Li, Qing},
  journal={arXiv preprint arXiv:2412.15606},
  year={2024}
}

@article{shi2025legalreasoner,
  title={LegalReasoner: Step-wised Verification-Correction for Legal Judgment Reasoning},
  author={Shi, Weijie and Zhu, Han and Ji, Jiaming and Li, Mengze and Zhang, Jipeng and Zhang, Ruiyuan and Zhu, Jia and Xu, Jiajie and Han, Sirui and Guo, Yike},
  journal={arXiv preprint arXiv:2506.07443},
  year={2025}
}

@article{tan2025enhancing,
  title={Enhancing Logical Reasoning in Language Models via Symbolically-Guided Monte Carlo Process Supervision},
  author={Tan, Xingwei and Valentino, Marco and Akhter, Mahmud and Liakata, Maria and Aletras, Nikolaos},
  journal={arXiv preprint arXiv:2505.20415},
  year={2025}
}

@article{huang2022large,
  title={Large language models can self-improve},
  author={Huang, Jiaxin and Gu, Shixiang Shane and Hou, Le and Wu, Yuexin and Wang, Xuezhi and Yu, Hongkun and Han, Jiawei},
  journal={arXiv preprint arXiv:2210.11610},
  year={2022}
}

@article{smit2023should,
  title={Should we be going mad? a look at multi-agent debate strategies for llms},
  author={Smit, Andries and Duckworth, Paul and Grinsztajn, Nathan and Barrett, Thomas D and Pretorius, Arnu},
  journal={arXiv preprint arXiv:2311.17371},
  year={2023}
}

@article{liang2023encouraging,
  title={Encouraging divergent thinking in large language models through multi-agent debate},
  author={Liang, Tian and He, Zhiwei and Jiao, Wenxiang and Wang, Xing and Wang, Yan and Wang, Rui and Yang, Yujiu and Shi, Shuming and Tu, Zhaopeng},
  journal={arXiv preprint arXiv:2305.19118},
  year={2023}
}

@article{sun2025mm,
  title={Mm-verify: Enhancing multimodal reasoning with chain-of-thought verification},
  author={Sun, Linzhuang and Liang, Hao and Wei, Jingxuan and Yu, Bihui and Li, Tianpeng and Yang, Fan and Zhou, Zenan and Zhang, Wentao},
  journal={arXiv preprint arXiv:2502.13383},
  year={2025}
}

@article{ficek2025scoring,
  title={Scoring Verifiers: Evaluating Synthetic Verification for Code and Reasoning},
  author={Ficek, Aleksander and Majumdar, Somshubra and Noroozi, Vahid and Ginsburg, Boris},
  journal={arXiv preprint arXiv:2502.13820},
  year={2025}
}

@article{lee2025revise,
  title={Revise: Learning to refine at test-time via intrinsic self-verification},
  author={Lee, Hyunseok and Oh, Seunghyuk and Kim, Jaehyung and Shin, Jinwoo and Tack, Jihoon},
  journal={arXiv preprint arXiv:2502.14565},
  year={2025}
}

@article{hendrycks2020measuring,
  title={Measuring massive multitask language understanding},
  author={Hendrycks, Dan and Burns, Collin and Basart, Steven and Zou, Andy and Mazeika, Mantas and Song, Dawn and Steinhardt, Jacob},
  journal={arXiv preprint arXiv:2009.03300},
  year={2020}
}

@inproceedings{rein2024gpqa,
  title={Gpqa: A graduate-level google-proof q\&a benchmark},
  author={Rein, David and Hou, Betty Li and Stickland, Asa Cooper and Petty, Jackson and Pang, Richard Yuanzhe and Dirani, Julien and Michael, Julian and Bowman, Samuel R},
  booktitle={First Conference on Language Modeling},
  year={2024}
}

@article{cobbe2021training,
  title={Training verifiers to solve math word problems},
  author={Cobbe, Karl and Kosaraju, Vineet and Bavarian, Mohammad and Chen, Mark and Jun, Heewoo and Kaiser, Lukasz and Plappert, Matthias and Tworek, Jerry and Hilton, Jacob and Nakano, Reiichiro and others},
  journal={arXiv preprint arXiv:2110.14168},
  year={2021}
}

@article{xiong2023can,
  title={Can llms express their uncertainty? an empirical evaluation of confidence elicitation in llms},
  author={Xiong, Miao and Hu, Zhiyuan and Lu, Xinyang and Li, Yifei and Fu, Jie and He, Junxian and Hooi, Bryan},
  journal={arXiv preprint arXiv:2306.13063},
  year={2023}
}

@inproceedings{liu2025breaking,
  title={Breaking mental set to improve reasoning through diverse multi-agent debate},
  author={Liu, Yexiang and Cao, Jie and Li, Zekun and He, Ran and Tan, Tieniu},
  booktitle={The Thirteenth International Conference on Learning Representations},
  year={2025}
}

@article{wang2022self,
  title={Self-consistency improves chain of thought reasoning in language models},
  author={Wang, Xuezhi and Wei, Jason and Schuurmans, Dale and Le, Quoc and Chi, Ed and Narang, Sharan and Chowdhery, Aakanksha and Zhou, Denny},
  journal={arXiv preprint arXiv:2203.11171},
  year={2022}
}

% \newpage
\appendix
\section{Limitations}
\label{limitations}
Although MARS achieved strong performance in both accuracy and efficiency, there are still several challenges remaining to be solved. (1) Confidence estimation. The reviewer agents output a confidence score, which helps the meta-reviewer to estimate the reliability of the review comments. In this paper, the confidence score is calibrated by a token probability-based uncertainty estimation. However, LLM uncertainty estimation is not yet a definitively solved problem, and incorporating more advanced calibration methods from recent studies remains a promising direction for future work. (2) Over-correction. In MARS, the decision and suggestions of meta-reviewer will highly influence the final answer, especially when the meta-decision is \textit{reject}. One potential issue is that a correct initial response may be overturned if the meta-reviewer provides negative feedback, guiding the author to a wrong direction. Although we attempted to mitigate this by curating the feedback prompt template (the author is asked to insist on its initial answer when disagreeing with the meta-reviewer), we cannot guarantee to avoid it in our current setting. Future work should include how to design more effective mechanisms to prevent unnecessary revisions. (3) Additionally, our current experiments constrain the maximum number of revision / debate rounds to two for both MARS and MAD due to computational resource and time limits. Systematically investigating the impact of additional interaction rounds on both reasoning performance and token consumption remains an important direction for future work.

\section{MARS Algorithm}
The standard workflow of MARS is shown in Figure~\ref{alg:mars}.

\begin{algorithm}[h]
\caption{MARS algorithm}
\label{alg:mars}
\KwIn{input query $x$, author model $\mathcal{A}$, $m$ reviewer models $\{\mathcal{R}_j \mid j=1,\dots,m\}$, meta-reviewer model $\mathcal{M}$, maximum iterations $K$}
\KwOut{final answer $y^*$}

\textbf{Initialization Stage:} \\
$t, y \leftarrow \mathcal{A}(x)$ \tcp*{Generate thoughts and initial answer (Eq.~\ref{eq_author})}
$k \leftarrow 1$ \\
$decision \leftarrow \text{Reject}$ \\

\While{$decision == \text{Reject}$ \textbf{and} $k \le K$}{
    \textbf{Review Stage:} \\
    \For{$j=1$ \KwTo $m$}{
        $r_j \leftarrow \mathcal{R}_j(x, t, y)$ \tcp*{Reviewers output decision, confidence, and justification}
    }
    $r \leftarrow \{r_j \mid j=1,\dots,m\}$ \tcp*{Collect all reviewer comments}
    
    \textbf{Meta-Review Stage:} \\
    $decision, f \leftarrow \mathcal{M}(x, t, y, r)$ \tcp*{Meta-reviewer evaluates reviews to output final decision and feedback $f$}
    
    \If{$decision == \text{Accept}$}{
        \textbf{break} \tcp*{Terminate loop early if answer is accepted}
    }
    
    \textbf{Revision Stage:} \\
    $t, y \leftarrow \mathcal{A}(x, t, y, f)$ \tcp*{Author revises thoughts and answer based on feedback $f$}
    $k \leftarrow k + 1$
}

$y^* \leftarrow y$ \tcp*{Final answer is the most recent $y$}
\Return{$y^*$}
\end{algorithm}

\section{Baselines and Experiment Details}
\label{ex_details}
We compare MARS against the following baselines: 
\begin{itemize}
    \item \textbf{Chain-of-Thought (CoT).} All models in our experiments are enabled with CoT, which prompts step-by-step reasoning and serves as the most basic baseline.
    \item \textbf{Self-reflection.} This is a single-agent approach in which the model is prompted to evaluate and revise its own response. 
    \item \textbf{Self-consistency.} This approach exploits the inherent stochasticity of LLMs by sampling multiple reasoning paths for the same query and aggregating them through majority voting to produce the final answer.
    \item \textbf{Multi-agent Debate (MAD)} is a framework in which multiple agents engage in a round-table discussion, with each agent generating an answer by combining its own reasoning with the responses of others. All agents play equivalent roles in this process.
    \item \textbf{Diverse Multi-agent Debate (DMAD)} is a recent variant of the MAD framework. Unlike traditional MAD, where all agents uniformly employ Chain-of-Thought (CoT) reasoning, DMAD assigns distinct reasoning strategies to different agents to explicitly foster cognitive diversity.
\end{itemize}

In our experiments, we followed the standard manner of CoT by asking the model to think about the user query step-by-step. For a fair comparison, we adopt a consistent prompt design: all methods use CoT prompting for the initial reasoning process. For example, the initial reasoning step in self-reflection and each reasoning path in self-consistency share the same CoT prompt template. Reflection, debate, review, and feedback methods require distinct prompt styles due to their structural differences. Detailed prompt templates are provided in Appendix~\ref{all_prompts}.

We used the standard OpenAI SDK for API calls of all the models used in this work. All of the parameters are set as their default values. For fair comparison, we use the same number of agents in MARS and MAD. Both methods will run one update round after the initial round. While MARS is naturally extendable to multiple rounds of review and revision, we leave a systematic study of this setting to future work due to time and cost constraints.

For each dataset, we randomly sample 100 problems in each experiment, and report the averaged performance after 3 runs.

\section{Additional Experimental Results}
The averaged inference time per query of MARS and the compared baselines are shown in Table~\ref{tab:inference_time}.

\begin{table*}[t]
\caption{Comparison of average inference time (seconds) per query across datasets and models. The lowest inference time among multi-agent methods is \textbf{bolded} for each setting.}
\centering
\renewcommand{\arraystretch}{1.1}
\small % Adjusting font size for consistency with the main table
\begin{tabular}{l ccccc}
\toprule
Method & GPT-3.5 & GPT-4o-mini & Mixtral 8x7b & Mixtral 8x22b & Llama3-70b \\
\midrule
\multicolumn{6}{c}{\textbf{GPQA}} \\
CoT              & 1.58 & 7.63 & 16.05 & 3.92 & 7.61 \\
Self-reflection  & 2.44 & 11.45 & 38.08 & 6.00 & 13.24 \\
Self-consistency & 6.37 & 29.20 & 68.17 & 23.11 & 52.33 \\
MAD              & 17.04 & 85.67 & 79.33 & 36.83 & 104.25 \\
DMAD             & 12.77 & 84.06 & 128.67 & 33.78 & 102.78 \\
MARS             & 9.54 & 49.02 & 79.24 & 22.21 & 49.77 \\
\midrule
\multicolumn{6}{c}{\textbf{MMLU}} \\
CoT              & 2.07 & 3.48 & 20.38 & 2.29 & 2.73 \\
Self-reflection  & 4.28 & 5.42 & 29.82 & 3.90 & 6.86 \\
Self-consistency & 4.93 & 14.55 & 47.76 & 14.44 & 29.60 \\
MAD              & 11.38 & 41.58 & 77.31 & 17.39 & 51.78 \\
DMAD             & 10.67 & 42.96 & 105.87 & 23.89 & 64.03 \\
MARS             & 7.61 & 28.14 & 55.91 & 16.15 & 36.49 \\
\midrule
\multicolumn{6}{c}{\textbf{GSM8K}} \\
CoT              & 1.67 & 3.48 & 4.32 & 2.50 & 2.24 \\
Self-reflection  & 2.63 & 5.42 & 26.84 & 3.46 & 4.31 \\
Self-consistency & 5.97 & 18.53 & 47.78 & 14.63 & 26.35 \\
MAD              & 11.58 & 39.69 & 79.21 & 23.55 & 28.67 \\
DMAD             & 14.47 & 46.86 & 54.61 & 20.31 & 70.35 \\
MARS             & 7.17 & 23.16 & 63.15 & 16.23 & 27.53 \\
\bottomrule
\end{tabular}
\label{tab:inference_time}
\end{table*}

\section{Prompt Templates}
\label{all_prompts}
In this section, we provide prompt templates for all the methods used in this paper, including both MARS and the compared baselines. 

\subsection{Prompt Templates for MARS}
\subsubsection{Author Agent}
\label{author_prompt}
The author agent is prompted to give its initial response using chain-of-thoughts. The following prompt shows that the author receives the user query, generates intermediate thoughts on the problem, and then the final answer. Specifically, the answer format shown here is for QA problems. When applied to math datasets, the format could be changed to 

``Answer: [the final numerical answer]''\\
``Your final answer must be a single numerical number at the end of the response.''
\begin{lstlisting}[basicstyle=\ttfamily\small, frame=single, breaklines=True]
You are an assistant. Please help to solve the following problem:
{user_query}

Give your thoughts about the question and the final answer in the following format:
Thoughts: [your thoughts with immediate results]
Answer: [the final single capital letter answer in the form (X). X is chosen from [A,B,C,D]]

Your final answer must be a single capital letter in the form (X) 
at the end of the response.
\end{lstlisting}

\subsubsection{Reviewer Agent}
\label{reviewer_prompt}
\begin{lstlisting}[basicstyle=\ttfamily\small, frame=single, breaklines=true]
You are an objective reviewer. The author has submitted an answer to a problem.
Your goal is to check for correctness, NOT to find faults that don't exist.

Criteria: Check for factual correctness and reasoning. Verify why the other options are wrong.

Question: {user_query}

Author's Answer: {author_response}

---
OUTPUT FORMAT:
My Independent Analysis: [Analyze the question and evaluate each option A, B, C, D]
My Selected Option: [e.g., (A)]
Comparison: [Does your option match the Author's?]
Decision: [right | wrong] (Vote 'right' if the Author's option matches yours)
Confidence: [1-5]
Justification: [Explain why the Author's reasoning is correct or incorrect]
---
\end{lstlisting}

\subsubsection{Confidence Calibration}
\label{confidence}
\begin{lstlisting}[basicstyle=\ttfamily\small, frame=single, breaklines=true]
NOTE ON RELIABILITY SCORES:
- Each Reviewer provides a 'Reliability Score' (0.0 to 1.0) based on their internal model uncertainty.
- High Score (>0.8): The Reviewer is mathematically confident. Trust their specific calculations/facts.
- Low Score (<0.6): The Reviewer is uncertain or confused. You should be skeptical of their critique, especially if they disagree with a High-Score reviewer.
\end{lstlisting}

\subsubsection{Meta-Reviewer Agent}
\label{meta_prompt}
\begin{lstlisting}[basicstyle=\ttfamily\small, frame=single, breaklines=true]
You are the meta-reviewer. The author has submitted an answer.

Question: {user_query}
Answer: {author_response}

You must decide whether the answer is correct based on both your own 
knowledge and the reviewers' comments below:

--- Reviewer Feedback ---
{combined_reviews}

Do not only rely on the reviewers, you must also think by yourself.

Provide your conclusion in the following format:

Decision: [right | wrong]
Justification: [reasons of your decision]
Suggestions: [your suggestions for updating the answer, only needed
              when decision is wrong]
Answer: [your recommended answer]
\end{lstlisting}

\subsubsection{Feedback to Author}
\label{feedback_prompt}
To avoid over-correction, the author is prompted to think carefully before accepting the meta-reviewer's suggestions. The author will only start to revise its initial answer when it strongly agrees with the meta-reviewer. After the decision is made, it will move on to update its answer following the feedback.
\begin{lstlisting}[basicstyle=\ttfamily\small, frame=single, breaklines=true]
Your answer was reviewed and marked as incorrect by the meta-reviewer.

--- Meta-reviewer Feedback ---
{meta_decision}

If you strongly agree with the meta-reviewer's suggestions, revise your
answer accordingly. If you disagree, insist on your initial answer and 
repeat it. Do not always trust the meta-reviewer -- you must think for 
yourself whether to trust the suggestions.

Make sure to state your reasoning and final answer in this format:

Reasons: [your reasons of accepting or rejecting the suggestions]
Thoughts: [your new step-by-step thoughts on the problem after 
           considering the suggestions]
Answer: [the final single capital letter answer in the form (X). 
         X is chosen from [A,B,C,D]]
\end{lstlisting}

\subsection{Prompt Templates for baselines}

\label{baseline_prompt}
In this section, we show the prompt templates for the baselines. For a fair comparison, the same CoT template is applied to the initial input of all the techniques. Therefore, we did not include the prompts for self-consistency here since it just applies the same prompt several times. For self-reflection, we provide the prompt for the reflection step. For MAD, we provide both the prompts for the initial response and the debate process for each agent.
\subsubsection{Self-Reflection Prompts}
\begin{lstlisting}[basicstyle=\ttfamily\small, frame=single, breaklines=true]
You wrote the following response to a problem:

Question: {user_query}
Answer: {response}

Carefully review your own answer. Are there any mistakes or thoughts not grounded in the given problem or known facts?
- If yes, explain the problems and revise your answer accordingly.
- If not, confirm and repeat your initial answer.

Your final response must follow this format:
Mistakes (if any):
Answer: [the final single capital letter answer in the form (X). 
         X is chosen from [A,B,C,D]]
\end{lstlisting}

\subsubsection{Debate Prompt}
\begin{lstlisting}[basicstyle=\ttfamily\small, frame=single, breaklines=true]
You are an assistant. Please help to solve the following problem:
{user_query}

Give your thoughts about the computation steps and the final answer in the following format:
Thoughts: [your step-by-step thinking process with immediate results]
Answer: [the final single capital letter answer in the form (X). 
         X is chosen from [A,B,C,D]]

Your final answer must be a single capital letter at the end of the response.
\end{lstlisting}

\begin{lstlisting}[basicstyle=\ttfamily\small, frame=single, breaklines=true]
These are the solutions to the problem from other agents:

One agent solution: ```{response_1}```
One agent solution: ```{response_2}```
...

Using the solutions from other agents as additional information, 
can you provide your final answer to the problem?

Make sure to state your thoughts and new answer with this format:
Thoughts: [your step-by-step thinking process]
Answer: [the final capital letter answer]

Your final answer must be a single capital letter at the end of the response.
\end{lstlisting}

\subsection{Reviewer Personas}
\label{persona_prompt}
Conservative Persona
\begin{lstlisting}[basicstyle=\ttfamily\small, frame=single, breaklines=true]
Your decision should be conservative.

Goal:
Minimize false alarms. Only label 'wrong' when there is a decisive, 
demonstrable error that changes the final answer.

Heuristics:
- Require at least TWO independent checks (e.g., recomputation + boundary/unit check) before declaring 'wrong'.
- Minor arithmetic slips that do not affect the final answer => prefer 'right' with notes in Justification.
- Missing small steps != 'wrong' if the step is standard and the result follows.

Confidence:
- 5 only if two checks agree
- 4 if one check is strong
- <=3 if any uncertainty remains
\end{lstlisting}

Aggressive Persona
\begin{lstlisting}[basicstyle=\ttfamily\small, frame=single, breaklines=true]
Your decision should be aggressive.

Goal:
Minimize false acceptances. Label 'wrong' when any essential step is missing, unjustified, or numerically inconsistent.

Heuristics:
- Penalize ambiguous leaps, missing assumptions/units, and unverified claims even if the final number looks plausible.
- If any step is unsupported or contradicts problem constraints, prefer 'wrong'.
- If correct, still list at least two vulnerabilities the author should fix.

Confidence:
- 5 if multiple issues are found or if cross-checks clearly contradict
- 4 when evidence is strong but not exhaustive
- <=3 if partial doubt
\end{lstlisting}

\section{LLM Usage Statement}
Large language models (LLMs) were used solely as general-purpose assistive tools for this work. Their usage was limited to grammar checking, language polishing, and LaTeX table formatting. LLMs were not involved in research ideation, methodological design, data analysis, or substantive writing of the manuscript.

\section{Additional Case Studies}
\label{append_case}

\begin{table}[h!]
\centering
\renewcommand{\arraystretch}{1.25}
\setlength{\tabcolsep}{6pt}
\begin{tabular}{|l|p{12cm}|}
\hline
\textbf{Question} & A photon $h\nu$ propagates in the Oz direction of the laboratory frame (R). It collides elastically with a particle $m$ at rest in (R). We denote $\alpha=\frac{h\nu}{mc^{2}}$, and $v=\beta c$ the velocity of the COM frame (R') of the system photon-particle relative to (R). What is the expression of $\beta$? \newline
Options: \newline 
(A) $\alpha\times\left(\frac{1}{1+\alpha}\right)$ \newline 
(B) $\alpha\times\left(\frac{1}{\sqrt{1+\alpha}}\right)$ \newline 
(C) $\alpha\times\left(\frac{1}{1-\alpha}\right)$ \newline 
(D) $\alpha\times\left(\frac{1}{\sqrt{1-\alpha}}\right)$ \\
\hline
\textbf{Author Response} & Thoughts: The problem requires determining the velocity $\beta c$ of the COM frame. Using conservation of momentum and energy, the motion must account for the photon's energy and the particle's rest mass. Relativistic mechanics often involve a correction factor like $\sqrt{1 \pm \alpha}$. Examining the options, B seems most suitable based on common relativistic calculations involving square roots. \newline 
Answer: (B) \\
\hline
\textbf{Reviewer 1} & Decision: wrong \newline 
Confidence: 5 \newline 
Justification: By conservation of energy, $E_{\text{total}} = h\nu + mc^2 = (1 + \alpha)mc^2$. By conservation of momentum, $p_{\text{total}} = \frac{h\nu}{c}$. The COM velocity is given by $\beta c = \frac{p_{\text{total}} c}{E_{\text{total}}}$. This simplifies to $\beta = \frac{\alpha}{1 + \alpha}$. The author incorrectly assumed a square root is necessary. 
\\
\hline
\textbf{Reviewer 2} & Decision: wrong \newline 
Confidence: 4 \newline 
Justification: The author correctly discussed conservation principles but arrived at the wrong numerical expression. The correct formulation is based directly on the conservation of momentum and energy, yielding $\beta = \frac{\alpha}{1+\alpha}$ without a square root. 
\\
\hline
\textbf{Meta-Review} & Decision: wrong \newline 
Justification: Both Reviewers analyzed the problem independently and showed detailed steps relying on conservation equations, verifying that $\beta$ depends on $1 + \alpha$. The author's inclusion of a square root deviated from the correct formula and added unnecessary complexity. \newline 
Suggestions: Revisit the momentum conservation and relate it correctly to the total energy without introducing arbitrary square root factors. 
\\
\hline
\textbf{Author Rebuttal} & Thoughts: Upon reevaluating, I recognize the importance of correctly applying conservation of momentum and energy. The total energy before the collision is $(1+\alpha)mc^2$ and the momentum relates to the COM frame, leading directly to $\beta = \frac{\alpha}{1+\alpha}$. The confusion in my initial answer stemmed from the improper inclusion of a square root term, which is not necessary in this fundamental particle-photon interaction setup. \newline 
Answer: (A) \\
\hline
\end{tabular}
\caption{Case study from GPQA: The MARS framework successfully corrects a heuristic hallucination by relying on rigorous physics derivations provided by the reviewer agents.}
\label{tab:case_study_physics}
\end{table}

\begin{table*}[t!]
\centering
\renewcommand{\arraystretch}{1.3}
\setlength{\tabcolsep}{6pt}
\begin{tabular}{|l|p{12cm}|}
\hline
\textbf{Question} & Consider a system of 3 spin-1/2 particles with a Hamiltonian $H = A(\mathbf{S}_1\cdot\mathbf{S}_2 + \mathbf{S}_1\cdot\mathbf{S}_3 + \mathbf{S}_2\cdot\mathbf{S}_3)$. The state is $\psi=\frac{1}{\sqrt{3}}(|\uparrow\uparrow\downarrow\rangle + |\uparrow\downarrow\uparrow\rangle + |\downarrow\uparrow\uparrow\rangle)$. Is this an eigenfunction of $H$? \newline
(A) It is an eigenstate of $H$, with eigenvalue $\frac{3}{4}A\hbar^2$ \newline 
(B) It is not an eigenstate of $H$, with expectation value $\frac{2}{3}A\hbar^2$ \newline 
(C) It is an eigenstate of $H$, with eigenvalue $\frac{2}{3}A\hbar^2$ \newline 
(D) It is not an eigenstate of $H$, with expectation value $\frac{3}{4}A\hbar^2$ \\
\hline
\textbf{Author Response} & Thoughts: The state $\psi$ is a symmetric superposition of spin configurations of three particles. Given that it represents one of the symmetric combinations, it tends to suggest a mixed state rather than a defined eigenvalue. \newline 
Answer: (D) \\
\hline
\textbf{Reviewer 1} & Decision: wrong \newline 
Confidence: 5 \newline 
Justification: $\psi$ is a symmetric combination of three particles. Any pair of two spins in this symmetric state operates as a triplet. For a triplet state, $\mathbf{S}_i \cdot \mathbf{S}_j = \frac{1}{4}\hbar^2$. The Hamiltonian has three such pairs, so $H\psi = 3 \times A(\frac{1}{4}\hbar^2)\psi = \frac{3}{4}A\hbar^2\psi$. It is indeed an eigenstate.
\\
\hline
\textbf{Reviewer 2} & Decision: wrong \newline 
Confidence: 4 \newline 
Justification: The state lacks the sole triplet representation, hence is not an eigenstate. Computing the expectation value yields $\frac{2}{3}A\hbar^2$. 
\\
\hline
\textbf{Meta-Review} & Decision: wrong \newline 
Justification: Reviewer 1 provides a thorough and robust analysis verifying that $\psi$ is an eigenstate due to its symmetric properties, correctly deriving the energy $\frac{3}{4}A\hbar^2$. Reviewer 2's expectation calculation is contradicted by this proof. \newline 
Suggestions: The author should calculate the action of $H$ on the stated $\psi$ more explicitly, as demonstrated by Reviewer 1.
\\
\hline
\textbf{Author Rebuttal} & Thoughts: After reevaluating, it is clear that $\psi$ forms a fully symmetric combination. The Hamiltonian acts on pairs of spin operators. Due to this symmetry, $\psi$ is an eigenstate of $H$, and summing the triplet pair interactions leads to the specific eigenvalue $\frac{3}{4}A\hbar^2$. \newline 
Answer: (A) \\
\hline
\end{tabular}
\caption{Case study from GPQA: The MARS Meta-Reviewer successfully adjudicates between conflicting reviewers, selecting the rigorous quantum mechanical derivation over flawed reasoning.}
\label{tab:case_study_quantum}
\end{table*}

\end{document}